\documentclass[runningheads]{llncs}
\usepackage{graphicx}
\usepackage{epstopdf}
\usepackage{amsmath,amssymb}
\usepackage{algorithm}
\usepackage{algpseudocode}
\usepackage{amsmath}
\usepackage{graphics}
\usepackage{epsfig}
\usepackage{multirow}
\usepackage{array}
\usepackage{subfigure}
\usepackage{booktabs}
\begin{document}

\title{Cascaded Pyramid Mining Network for Weakly Supervised Temporal Action Localization} 
\titlerunning{Cascaded Pyramid Mining Network} 


\author{Haisheng Su \and
Xu Zhao\thanks{Corresponding author. This research is supported by the funding from NSFC programs (61673269, 61273285, U1764264).} \and
Tianwei Lin}
%

\authorrunning{Haisheng Su et al.} 


\institute{Department of Automation, Shanghai Jiao Tong University, China 
\email{\{suhaisheng,zhaoxu,wzmsltw\}@sjtu.edu.cn}}

\maketitle

\begin{abstract}
Weakly supervised temporal action localization, which aims at temporally locating action instances in untrimmed videos using only video-level class labels during training, is an important yet challenging problem in video analysis. Many current methods adopt the ``\emph{localization by classification}'' framework: first do video classification, then locate temporal area contributing to the results most. However, this framework fails to locate the entire action instances and gives little consideration to the local context. In this paper, we present a novel architecture called \textbf{Cascaded Pyramid Mining Network (CPMN)} to address these issues using two effective modules. First, to discover the entire temporal interval of specific action, we design a two-stage cascaded module with proposed \textbf{Online Adversarial Erasing (OAE)} mechanism, where new and complementary regions are mined through feeding the erased feature maps of discovered regions back to the system. Second, to exploit hierarchical contextual information in videos and reduce missing detections, we design a pyramid module which produces a scale-invariant attention map through combining the feature maps from different levels. Final, we aggregate the results of two modules to perform action localization via locating high score areas in temporal \textbf{Class Activation Sequence (CAS)}. Extensive experiments conducted on THUMOS14 and ActivityNet-1.3 datasets demonstrate the effectiveness of our method.

\end{abstract}
\section{Introduction}
Due to the rapid development of computer vision along with the increasing amount of videos, many breakthroughs have been observed on video content analysis in recent years. Videos from realistic scenarios are often complex, which may contain multiple action instances of different categories with varied lengths. This problem leads to a challenging task: temporal action localization, which requires to not only handle the category classification of untrimmed videos but also determine the temporal boundaries of action instances. Nevertheless, it implies the huge amounts of temporal annotations for training an action localization model, which are more labor-intensive to obtain than video-level class labels. 

Contrary to the fully supervised counterparts, Weakly Supervised Temporal Action Localization (WSTAL) task learns TAL using only video-level class labels, which can be regarded as a temporal version of Weakly Supervised Object Detection (WSOD) in image. A popular series of models in WSOD generate Class Activation Maps (CAMs) \cite{B.Zhou} to highlight the discriminative object regions contributing to the classification results most. Inspired by \cite{B.Zhou}, recently many WSTAL works generate the Class Activation Sequence (CAS) to locate the action instances in temporal domain. However, many drawbacks have been observed in this ``\emph{localization by classification}'' mechanism: (1) the CAS fails to generate dense detections of target actions, causing many missing detections; (2) the classification network usually leverages features of discriminative rather than entire regions for recognition, failing to handle the action instances with varied lengths;
(3) some true negative regions are falsely activated, which is mainly due to the action classifier realizes the recognition task based on a global knowledge of the video, resulting in inevitably neglecting the local details. 

\begin{figure}[t]
	\setlength{\abovecaptionskip}{-0.3cm} 
	\setlength{\belowcaptionskip}{-0.5cm}
	\begin{center}
		\begin{minipage}[b]{1.0\linewidth}
			\centering
			\centerline{\includegraphics[height=7cm]{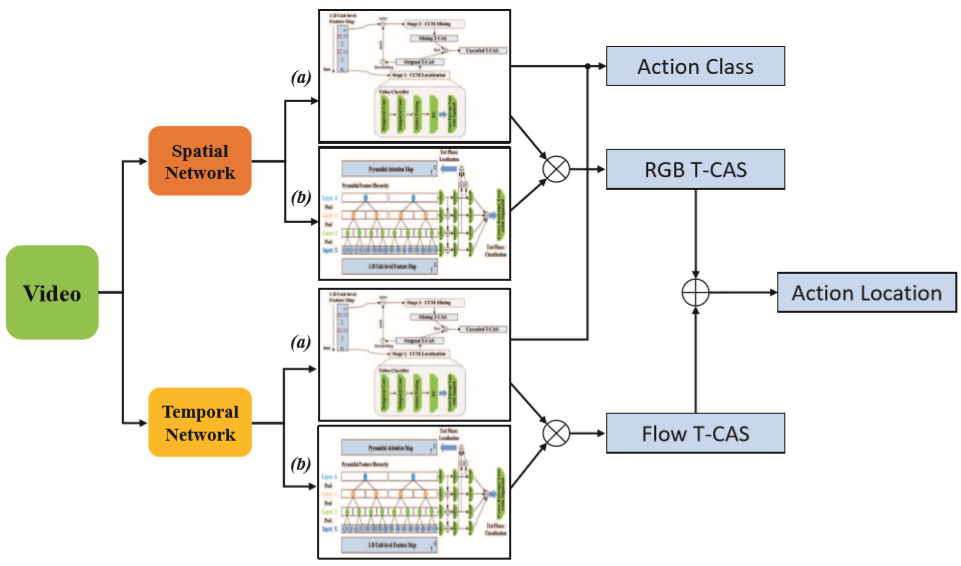}}
			\medskip
		\end{minipage}
	\end{center}
	\caption{Overview of our approach. Two-stream network is used to encode visual features for our algorithm to perform action classification and temporal action localization concurrently. The (a) Cascaded Classification Module (CCM) with Online Adversarial Erasing (OAE) method and the (b) Pyramid Attention Module (PAM) are proposed to compute attention-based cascaded Temporal Class Activation Sequence (T-CAS) from the two streams separately, which can be employed to locate the entire regions of specific actions in temporal domain with high accuracy.}
	\label{fig_overview}
\end{figure}

To address these issues and generate high quality detections, we propose the Cascaded Pyramid Mining Network (CPMN), which adopts two effective modules to mine entire regions of target actions and remove the false positive regions respectively. Specifically, CPMN generates detections in three steps. \textbf{First}, CPMN adapts two classifiers with different input feature maps to discover discriminative regions separately, and the input feature maps of the second classifier are erased with the guidance of the CAS from the first one. \textbf{Second}, CPMN combines the discriminative regions discovered by the two classifiers to form the entire detections. \textbf{Final}, taking full advantage of hierarchical contextual representations, CPMN generates a scale-invariant attention map to correct the false positive regions and reduce the missing detections. These pyramidal feature representations offer ``\emph{local to global}" context information for better evaluation. The overview of our algorithm is illustrated in Fig. 1.

To sum up, the main contributions of our work are three-fold:

(1) We propose a new architecture (CPMN) for weakly supervised temporal action localization in untrimmed videos, where entire temporal regions of action instances are located with less missing detections.

(2) We introduce an Online Adversarial Erasing (OAE) method to discover entire regions of target actions using two cascaded classifiers with different input feature representations, and explicitly handle the action instances with varied lengths by exploiting hierarchical contextual information.

(3) Extensive experiments demonstrate that our method achieves the state-of-the-art performance on both THUMOS14 and ActivityNet-1.3 datasets.

\section{Related Work}
\noindent
{\bf Action recognition.} Action recognition has been widely studied in recent years, which aims to identify one or multiple action categories of a video. Earlier works mainly focus on hand-crafted feature engineering, such as improved Dense Trajectory (iDT) \cite{H.Wang,iDT}. With the development of convolutional neural networks, many deep learning based methods \cite{K.Simonyan,C.Feichtenhofer,TSN,D.Tran} have been applied to action recognition task and achieve convincing performance. Two-stream network \cite{K.Simonyan,C.Feichtenhofer,TSN} typically consists of two branches which  learn the appearance and motion information using RGB image and optical flow respectively. C3D network \cite{D.Tran} simultaneously captures appearance and motion features using a series of 3D convolutional layers. These action recognition models are usually adopted to extract frame or unit level visual representation in long and untrimmed videos.

\noindent
{\bf Weakly supervised object detection.} Weakly supervised object detection aims to locate the objects using only image-level labels. Current works mainly include bottom-up \cite{H.Bilen,P.Tang} and top-down \cite{B.Zhou,J.Zhang,Singh,Y.Wei} mechanisms. Proposals are first generated in \cite{H.Bilen,P.Tang} using selective search \cite{J.R.Uijlings} or edge boxes \cite{C.L.Zitnick}, which are further classified and the classification results are merged to match the image labels. 
Zhou et al. \cite{B.Zhou} and Zhang et al. \cite{J.Zhang} aim to find out the relationship between the neural responses of image regions and classification results, and then locate top activations area as detections. Singh et al. \cite{Singh} propose to improve the localization map by randomly hiding some regions during training, so as to force the network to look for other discriminative parts. However, without effective guidance, this attempt is blind and inefficient. Recently, Wei et al. \cite{Y.Wei} employ Adversarial Erasing (AE) approach to discover more object regions in images by training classification network repeatedly, with discriminative regions erasing of different degrees, which is somewhat impractical and time-consuming. Our work differs from these methods in designing an Online Adversarial Erasing (OAE) approach which only needs to train a network for entire regions mining.

\noindent
{\bf Weakly supervised temporal action Localization.} Action localization in temporal domain \cite{T.Lin,H.Su,H.Xu,Yuwei.Chao} is similar to object detection in spatial domain \cite{RCNN,Dollar,maskrcnn,yolo,yolo9000}, as well as the case under weak supervision. WSTAL aims to locate action instances in untrimmed videos including both temporal boundaries and action categories while relying on video-level class label only. Based on the idea proposed in \cite{H.Bilen}, Wang et al. \cite{L.Wang} formulate this task as a proposal-based classification problem, where temporal proposals are extracted with the priors of action shot. However, the use of softmax function across proposals blocks it from distinguishing multiple action instances. Singh et al. \cite{Singh} hide temporal regions to force attention learning. However, it's not applicable owing to the complexity and varied lengths of videos. In our work, the Pyramid Attention Module (PAM) is proposed to hierarchically classify the videos from local to global, thus the pyramidal attention map is generated by combining feature maps from different levels, which can be scale-invariant to the action instances. 

\section{Our Approach}
\subsection{Problem Definition}
We denote an untrimmed video as $ X_{v} = \{ x_{t}\}_{t=1}^{l_{v}}, $ where $ l_{v} $ is the number of frames and $ x_{t} $ is the $ t $-th frame in $ X_{v}. $ Each video $ X_{v} $ is annotated with a set of temporal action instances $ \Phi_{v} = \{ \phi_{n} = (t_{n}^{s}, t_{n}^{e}, \varphi_{n})\}_{n=1}^{N_{v}}, $ where $ N_{v} $ is the number of temporal action instances in $ X_{v} $, and $ t_{n}^{s}, t_{n}^{e}, \varphi_{n} $ are starting time, ending time and category of instance $ \phi_{n} $ respectively, where $ \varphi_{n}\in\{1,...,C\} $ and $ C $ is the number of action categories. During training phase, only the video-level action label set $ \psi_{v} = \{\varphi_{n}\}_{n=1}^{N_{v}}$ is given, and during test phase, $ \Phi_{v} $ need to be predicted.

\subsection{Video Features Encoding}
To apply CPMN, first feature representations need to be extracted to describe visual content of the input video in our work. UntrimmedNet \cite{L.Wang} is employed as visual encoder, since this kind of architecture using multiple two-stream networks has shown great performance and becomes a prevalent practice adopted in action recognition and temporal action localization tasks. 

Given a video containing $ l_{v} $ frames, we use video unit as the basic processing unit in our framework for computational efficiency. Hence the video is divided into $ l_{v}/n_{u} $ consecutive video units without overlap, where $ n_{u} $ is the frame number of a unit. Then we compose a unit sequence $ U = \{u_{j}\}_{j=1}^{l_{u}} $ from the video $ X_{v} $, where $ l_{u} $ is the number of units. A video unit can be represented as $ u_{j} = \{x_{t}\}_{t=f_{s}}^{f_s+n_{u}} $, where $ f_{s} $ is the starting frame and  $ f_s+n_{u} $ is the ending frame. Each unit is fed to the pre-trained visual encoder to extract representation. Concretely, the center RGB frame inside a unit is processed by spatial network and stacked optical flow derived around the center frame is processed by temporal network, then we concatenate output scores of UntrimmedNet \cite{L.Wang} in fc-action layer to form the feature vector $ f_{u_{j}} = \{f_{S,u_{j}}, f_{T,u_{j}} \}$, where $ f_{S,u_{j}} $ and $ f_{T,u_{j}} $ are output score vector of spatial and temporal network respectively with length $ G $. Final, the unit-level feature sequence $ F = \{f_{u_{j}}\}_{j=1}^{l_{u}} $ is used as input of CPMN.
               
\subsection{Cascaded Pyramid Mining Network}
To generate high quality detections with accurate temporal regions under weak supervision, we propose a multi-stage framework to achieve this goal. In CPMN, we first design a module to discover the entire action regions in a cascaded way. Then we introduce another module to combine the temporal feature maps from pyramidal feature hierarchy for prediction, making it possible to handle action instances with varied lengths. Final, we fuse the results from these two modules for action localization in temporal domain.

\noindent
{\bf Network architecture.} The architecture consists of three sub-modules: cascaded classification module, pyramid attention module and temporal action localization module. \textit{Cascaded classification module} is a two-stage model which includes two classifiers as shown in Fig. 2, aiming to mine different but complementary regions of target action in the video through a cascaded manner. \textit{Pyramid attention module} is proposed to generate the class probability of each input unit feature, through classifying the input feature sequence with hierarchical resolutions separately. The architecture of this module is illustrated in Fig. 3. Final, \textit{temporal action localization module} fuses the cascaded localization sequence and the pyramidal attention map to make it more accurate.

\noindent
{\bf Cascaded classification module.} The goal of this module is to locate the entire regions of target actions in the video, where two cascaded classifiers are needed. As shown in Fig. 2, the Cascaded Classification Module (CCM) contains two separate classifiers with the same structure. In each stage (i.e. the localization stage and the mining stage), the classifier handles the input unit-level features and adopts two 1-D temporal convolutional layers followed by a global average pooling layer to get the video-level representation, which is then passed through a fully connected (FC) layer and a Sigmoid layer for video classification. The two convolutional layers have the same configurations: kernel size 3, stride 1 and 512 filters with ReLU activation. Denote the last convolutional features and the averaged feature representation as $ \textbf{\textit{Z}} \in {\rm I\!R}^{T\times K} $ and $ \overline{\textbf{\textit{Z}}} = \{\frac{\sum _{t}Z_{t}(k)}{T}\}_{k=1}^{K} $ respectively, where $ T $ is the length of input feature sequence, $ K $ is the channel number and $ Z_{t}(k) $ is the $ k $-th feature map at time $ t $. Then based on the idea in \cite{B.Zhou}, we derive the 1-D Temporal Class Activation Sequence (T-CAS). We denote $ w^{c}(k) $ as the $ k $-th element of the weight matrix $ \textbf{\textit{W}} \in {\rm I\!R}^{K \times C} $ in the classification layer corresponding to class $ c $. The input to the Sigmoid layer for class $ c $ is
\begin{equation}
s^{c} = \sum_{k=1}^{K}  w^{c}(k)\overline{Z}(k)
= \sum_{k=1}^{K}  w^{c}(k)\sum_{t=1}^{T}Z_{t}(k) = \sum_{t=1}^{T}\sum_{k=1}^{K} w^{c}(k)Z_{t}(k),
\end{equation}
\begin{equation}
M_{t}^{c} = \sum_{k=1}^{K} w^{c}(k)Z_{t}(k),
\end{equation}
where $ M_{t}^{c} $ is denoted as T-CAS of class $ c $, which indicates the activations of each unit feature contributing to a specific class of the video.

\begin{figure}[t]
	\setlength{\abovecaptionskip}{-0.3cm}
	\setlength{\belowcaptionskip}{-0.3cm}
	\begin{center}
		\begin{minipage}[b]{1.0\linewidth}
			\centering
			\centerline{\includegraphics[height=6cm]{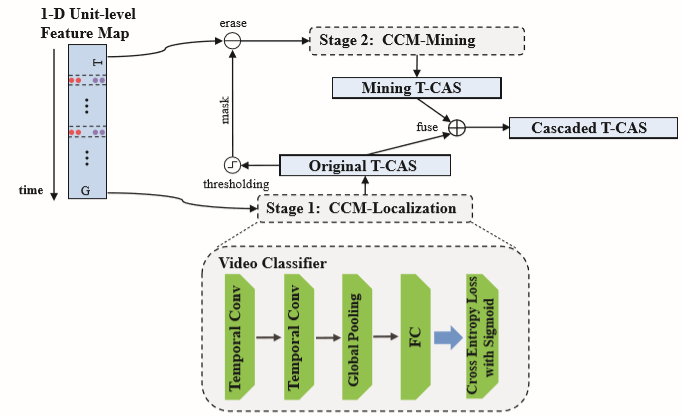}}
			\medskip
		\end{minipage}
	\end{center}
	\caption{Architecture of the cascaded classification module. The extracted unit-level feature representations are fed to two cascaded video classifiers for localization sequence inference individually. The two classifiers of the same structure share the input feature maps, and we erase the input features of discriminative regions highlighted by the first classifier, to drive the second classifier to discover more relevant regions of target actions in the video. Final, the two T-CASs are integrated for a better localization. }
	\label{fig_cascaded}
\end{figure}
\begin{figure}[t]
	\setlength{\abovecaptionskip}{-0.3cm} 
	\setlength{\belowcaptionskip}{-0.5cm} 
	\begin{center}
		\begin{minipage}[b]{1.0\linewidth}
			\centering
			\centerline{\includegraphics[height=5.2cm]{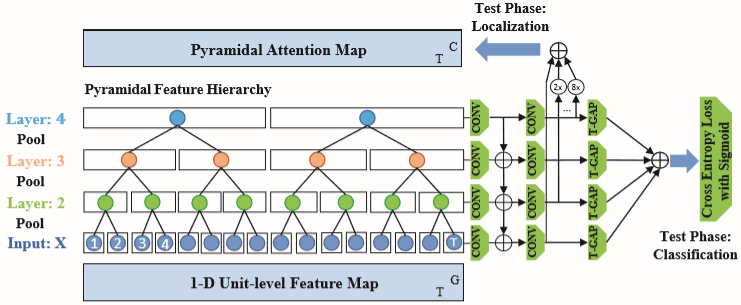}}	
			\medskip
		\end{minipage}
	\end{center}
	\caption{Architecture of the pyramid attention module used for generating the multi-scale attention map. The pyramid attention module consists of several branches based on feature representations of hierarchical resolutions, for the purpose of classifying the video from local to global. And a global average pooling layer is employed to encode the video-level prediction of each branch and then we aggregate the multiple prediction results from different levels. Final, the pyramidal attention map is constructed through combining the multi-scale feature maps in the prediction layers during test phase. }
	\label{fig_pyramid}
\end{figure}

Then we conduct a threshold on the T-CAS obtained in the first stage to generate a mask which represents the discriminative regions discovered by the first classifier, and the mask is used to erase the input features of the second stage for classification. Such an online adversarial erasing operation allows the second classifier to leverage features from other regions for supporting the video-level labels. Final, we integrate the two T-CASs, $ M^{c}(A)$ and $ M^{c}(B)$, which are generated in the two stages separately, to form the cascaded localization sequence $  M^{c}(Cas) $. Concretely, $  M^{c}_{t}(Cas) = max\{M^{c}_{t}(A), M^{c}_{t}(B)\} $,  where $ M^{c}_{t}(Cas) $ is the $ t $-th element in the cascaded action localization sequence of class $ c $.

\noindent
{\bf Pyramid attention module.} This module is designed to handle action instances with varied lengths. We achieve this goal in two steps. First, we semantically classify the feature representations with hierarchical temporal resolutions individually, aiming at processing the input unit-level feature sequence from local to global. Then we combine temporal feature maps in prediction layers from different levels to form the pyramidal attention map.

As shown in Fig. 3, we stack three 1-D max pooling layers on the input feature sequence, thus obtain the pyramidal feature hierarchy. After subsampling with a scaling step of 2 layer by layer, the feature sequence length of the $ l $-th layer is $ T_{l} = \frac{T}{2^{l-1}}$. Then for each level, we first conduct a temporal convolutional layer with kernel size 1, stride 1 and 512 filters to handle the feature sequence. Among these levels, we use lateral connections to exploit hierarchical context. Then we employ another convolutional layer to predict the classification scores of units associated with each feature map respectively. Each prediction layer consists of $ C $ filters with kernel size 1 and stride 1. And we continue to append a global average pooling layer on the label maps $ \textbf{\textit{K}}_{l} \in {\rm I\!R}^{T_{l}\times C}$ of each level separately to aggregate the video-level predictions. Final, we average among these prediction results to match the video-level class labels.

During test phase, we form the class heatmaps $ \textbf{H} = \{H^{c}\}_{c=1}^{C} $ by combining the output label maps in prediction layers from different levels. For example, with the coarser-resolution label map generated in $ l $-th layer, we upsample the temporal resolution by a factor of $ 2^{l-1} $ through repeating the score vector in temporal dimension. Then every class heatmap is normalized to [0, 1] as follows,
\begin{equation}
H^{c} = (H^{c} - min(H^{c}))/(max(H^{c}) - min(H^{c})),
\end{equation}
where $ H^{c} $ is the heatmap for class $ c $, which indicates the class probability of each unit feature semantically.

\begin{figure}[t]
	\setlength{\abovecaptionskip}{-0.3cm}
	\setlength{\belowcaptionskip}{-0.3cm}
	\begin{center}
		\begin{minipage}[b]{1.0\linewidth}
			\centering
			\centerline{\includegraphics[height=3.8cm]{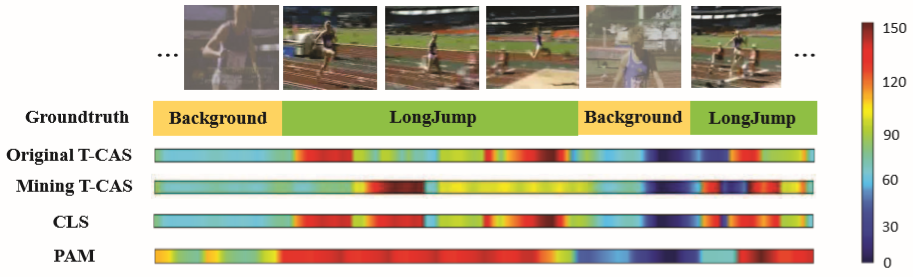}} 
			\medskip
		\end{minipage}
	\end{center}
	\caption{Illustration of the comparison of ground-truth temporal intervals, original T-CAS, mining T-CAS, cascaded localization sequence (CLS) and pyramidal attention map (PAM) for the \textit{LongJump} action class on THUMOS14.}
	\label{fig_loc}
\end{figure}

\noindent
{\bf Temporal action localization module.} The goal of this module is to fuse the cascaded action localization sequence and the pyramidal attention map obtained above for temporal action localization. As shown in Fig. 4, the CLS can identify more regions of target actions but some false positive regions are also activated by mistake. Since the class heatmap is generated using element-wise addition of label maps at multiple scales, the results are too smooth to indicate the accurate action boundaries. But they can provide important temporal priors to constrain the CLS. We let the class heatmap play the role of an attention map, which is used to correct the false positive regions and reduce missing detections. Then we fuse CLS and PAM to generate the attention-based cascaded T-CASs respectively as follows,
\begin{equation}
\setlength{\abovedisplayskip}{5pt} 
\Phi_{t,RGB}^{c} = sigmoid(M_{t,RGB}^{c}(Cas)) \cdot H_{t,RGB}^{c},
\setlength{\belowdisplayskip}{3pt}
\end{equation}
\begin{equation}
\setlength{\abovedisplayskip}{3pt} 
\Phi_{t,Flow}^{c} =sigmoid(M_{t,Flow}^{c}(Cas)) \cdot H_{t,Flow}^{c}.
\setlength{\belowdisplayskip}{3pt}
\end{equation}

Then for each action class, we conduct a threshold on the $ \Phi_{t,RGB}^{c} $ and $ \Phi_{t,Flow}^{c} $ separately, and different from the method \cite{B.Zhou} which only retains the bounding box that covers the largest connected components, we keep all connected units that pass the predefined threshold $ \theta_{TCAS} $ from each class and modality.

\subsection{Training of CPMN}
The training details of the cascaded classification module and the pyramid attention module in CPMN are introduced in this section.

\noindent
{\bf Cascaded classification module.} Given an input video $ X_{v} $, we form unit sequence $ U $ and extract corresponding feature sequence $ F $ with length $ l_{u} $. Since an untrimmed video usually comes up with extremely irrelevant frames with action instances only occupying small parts, we totally test three sampling methods to simplify the feature sequence for computational cost reduction: (1) uniform sampling: units are extracted with a regular interval $ \sigma $ from $ U $, thus the final unit sequence and feature sequence are $ U^{'} = \{u_{j}^{'}\}_{j=1}^{l_{u}^{'}} $ and $ F^{'} $ separately, where $ l_{u}^{'} = \frac{l_{u}}{\sigma} $; (2) sparse sampling: we first divide the video into $ P $ segments $ \{S_{1},S_{2},...,S_{P}\}$ with equal length, then during each training epoch, we randomly sample one unit from each segment to form the unit sequence of length $ P $; (3) shot-based sampling: considering the action structure, we sample the unit sequence $ U $ based on action shots, which are generated by shot boundary detector \cite{STPN}. Evaluation results of these sampling methods are shown in Section 4.3. 

After sampling, we construct the training data of each video as $ \Theta_{cas}(X_{v}) = \{U^{'}(X_{v}),F^{'}(X_{v}),\psi_{v}\} $ and taking it as input, the first classifier leverages the most discriminative regions for classification while the second classifier handles the erased feature sequence for entire regions mining. We test different erasing thresholds $ \zeta $ from 0.3 to 0.7 and the evaluation results are shown in Section 4.3. And we employ cross-entropy loss and $ l_{2} $ regularization loss as final loss function to train the two multi-label classifiers separately:
\begin{equation}
L_{cas} = \sum_{v=1}^{D}\psi_{v}\cdot log(y_{v}^{predict}) + \lambda\cdot L_{2}(\Xi_{cas}),
\end{equation}
where $ y_{v}^{predict} $ is the predicted class scores of the video and $ D $ is the number of training videos. $ \lambda $ balances the cross-entropy loss and $ l_{2} $ loss, and $ \Xi_{cas} $ is the cascaded classification model. Algorithm 1 illustrates the training procedure. 

\renewcommand{\algorithmicrequire}{ \textbf{Input:}}  
\renewcommand{\algorithmicensure}{ \textbf{Output:}}  
\begin{algorithm}[t]
	\caption{Training procedure of cascaded classification module.}
	\begin{algorithmic}[1]
		\Require Training data, $\Theta_{cas} = \{\Theta_{cas}(X_{v})\}_{v=1}^{D}$; threshold, $\zeta$; 
		\Ensure  Cascaded T-CAS, $M^{c}(Cas)$;
		\Function{Main}{ }
		\While{training is not convergent}
		\State $M^{c}(A), M^{c}(B) \gets$ \Call{Cas$\_$Train}{$\Theta_{cas}(X_{v})$, $\zeta$}	
		\EndWhile
		\State $M^{c}(Cas) \gets $ max($M^{c}_{t}(A), M^{c}_{t}(B)$)
		\State \Return{$M^{c}(Cas)$}
		\EndFunction  	
		\Function {Cas$\_$Train}{$\Theta_{cas}(X_{v})$, $\zeta$}
		\State  Extract the feature sequence $F^{'}(X_{v})$
		\State  Generate the original T-CAS $M^{c}(A) \gets$ infer($F^{'}(X_{v}), \psi_{v}$)
		\State  Generate the mask $ M^{c}(mask)  \gets$ $M^{c}(A)>\zeta$
		\State  Obtain the erased feature sequence $F^{'}_{erase}(X_{v}) \gets$ erase($F^{'}(X_{v}), M^{c}(mask)$)
		\State  Generate the mining T-CAS $M^{c}(B) \gets$ infer($F^{'}_{erase}(X_{v}), \psi_{v}$)
		\State \Return{$M^{c}(A), M^{c}(B)$}
		\EndFunction	
	\end{algorithmic}
\end{algorithm}

\noindent
{\bf Pyramid attention module.} The pyramid attention module is trained to handle the action instances with arbitrary intervals. Considering the maximum length $ d_{max} $ of ground-truth action instances in dataset, we slide windows with length $ T_{\omega} $ which can cover the $ d_{max} $. The training data of video $ X_{\upsilon} $ is constructed as $ \Theta_{mul}(X_{v}) = \{\Omega = \{U_{\omega},F_{\omega}\}_{\omega=1}^{N_{\omega}},\psi_{v} \}$, where $ N_{\omega}$ is number of windows. Taking a sliding window with corresponding feature sequence $ F_{\omega} $ as input, the pyramid attention module generates label maps with different lengths, then combine these maps in a bottom-up way and concatenate all windows results to form the video pyramidal attention maps. The multi-label cross-entropy loss function is also adopted to train the pyramid attention module.

\subsection{Prediction and Post-processing}
During prediction, we follow the same data preparation procedures of training phase to prepare the testing data, except for the following two changes: (1) in the cascaded classification module, we use uniform sampling method to sample the input feature sequence in order to increase the prediction speed and guarantee the stable results; (2) in the pyramid attention module, if the length of input video is shorter than $ T_{\omega} $, we will pad the input feature sequence to $ T_{\omega} $ so that there is at least one window for the multi-tower network to predict. Then given a video $ X_{v} $, we use CPMN to generate proposal set $ \Gamma = \{\tau_{n}\}_{n=1}^{N_{p}}$ based on the thresholding attention-based cascaded T-CAS of top-2 predicted classes \cite{L.Wang}, where $ N_{p} $ is the number of candidate proposals and we set the threshold $ \theta_{TCAS} $ as 20\% of the max value of the derived T-CAS. For each proposal denoted by $ [t_{start}, t_{end}] $, we first calculate the mean attention-based cascaded T-CAS among the temporal range of the proposal as $ p_{act} $:
\begin{equation}
p_{act} = \sum_{t=t_{start}}^{t_{end}}\frac{M_{t,RGB}^{c}(Cas) \cdot H_{t,RGB}^{c} + M_{t,Flow}^{c}(Cas) \cdot H_{t,Flow}^{c}}{2\cdot(t_{end}-t_{start}+1)} ,
\end{equation}
then we fuse $ p_{act} $ and the class scores $ p_{class} $ with multiplication to get the confidence score $ p_{conf} $:
\begin{equation}
p_{conf} = p_{act}\cdot p_{class}.
\end{equation}

Since we keep all connected units that pass the predefined threshold $ \theta_{TCAS} $ from each class and each modality as proposals, we may generate multiple predictions with different overlap. Then we conduct non-maximum suppression with predefined threshold $ \theta_{NMS} $ in these prediction results to remove the redundant predictions of confidence score $ p_{conf}$. Finally we get the prediction instances set $ \Gamma^{'} = \{\tau_{n}^{'}\}_{n=1}^{N_{p}^{'}}$, where $ N_{p}^{'} $ is the number of final prediction instances.

\section{Experiments}

\subsection{Dataset and Setup}
\noindent
{\bf Dataset.} {\bf ActivityNet-1.3} \cite{Anet} is a large-scale video dataset for action recognition and temporal action localization tasks used in the ActivityNet Challenge 2017 and 2018, which consists of 10,024 videos for training, 4,926 for validation and 5,044 for testing, with 200 action classes annotated. Each video is annotated with average 1.5 temporal action instances. {\bf THUMOS14} \cite{Jiang} dataset contains 1010 videos for validation and 1574 videos for testing with video-level labels of 101 action classes, while only a subset of 200 and 213 videos separately are temporally annotated among 20 classes. We train our model with the validation subset without using the temporal annotations. In this section, we compare our method with state-of-the-art methods on both ActivityNet-1.3 and THUMOS14.

\noindent
{\bf Evaluation metrics.} Following the conventions, we use mean Average Precision (mAP) as evaluation metric, where Average Precision (AP) is calculated on each class separately. We report mAP values at different Intersection over Union (IoU) thresholds. On ActivityNet-1.3, mAP with IoU thresholds $ \{0.5,0.75,0.95 \} $ and average mAP with IoU thresholds set $ \{0.5: 0.05: 0.95\} $ are used. On THUMOS14, mAP with IoU thresholds $ \{0.1,0.2,0.3,0.4,0.5\} $ are used.

\noindent
{\bf Implementation details.}  We use the UntrimmedNet \cite{L.Wang} and TSN \cite{TSN} for visual feature encoding of THUMOS14 and ActivityNet-1.3 separately, where ResNet network \cite{K.He} is used as spatial network and BN-Inception network \cite{S.Ioffe} is adopted as temporal network. The two visual encoders are both implemented using Caffe \cite{Y.Jia} and the TSN is pre-trained on ActivityNet-1.3. During feature extraction, the frame number of a unit $ n_{u} $ is set to 5 on THUMO14 and is set to 16 on ActivityNet-1.3. In CPMN, the cascaded classification module and the pyramid attention module are both implemented using TensorFlow \cite{Abadi}. On both datasets, the two modules are both trained with batch size 16 and learning rate 0.001 for 30 epochs, then 0.0001 for another 120 epochs. The erasing threshold $ \zeta $ used in CCM is 0.4 and the window size $ T_{\omega} $ used in the PAM is 64. Besides, the regularization term $ \lambda $ is 0.0025. For NMS, we set the threshold $ \theta_{NMS} $ to 0.3 on THUMO14 and 0.5 on ActivityNet-1.3 by empirical validation.

\subsection{Comparison with State-of-the-art Methods}
We first evaluate the overall results of our proposed framework for action localization and compare our method with several state-of-the-art approaches including both fully and weakly supervised methods. Table 1 illustrates the localization performance on the THUMOS14 dataset. We can observe that our proposed algorithm achieves better performance than the two existing weakly supervised methods, and is even competitive to some fully supervised approaches. For example, when the IoU threshold $ \alpha $ is 0.5, the mAP of our method is more than twice higher as the one of \cite{Singh}, which significantly convinces that our proposed online adversarial erasing strategy is more reasonable than randomly hiding. And the substantial performance gaining over the previous works under different IoU thresholds confirms the effectiveness of our CPMN. 

We also present our results on the validation set of the ActivityNet-1.3 dataset to further validate our localization performance. In ActivityNet-1.3, since the length of videos is not too long like THUMOS14, we directly resize the extracted feature sequence to length $ l_{\omega} $= 64 by linear interpolation. And we choose attention-based cascaded T-CAS of top-1 class \cite{Uts} as our detections. The evaluation results on ActivityNet-1.3 are shown in Table 2, from which we can see that our algorithm is generalized enough to this dataset, and significantly outperforms most fully supervised approaches with convincing performance. Meanwhile, we are the first to evaluate weakly supervised method on this dataset. 

\renewcommand{\multirowsetup}{\centering}
\setlength{\tabcolsep}{7pt}
\begin{table}[t]
	\begin{center}
		\caption{Comparison of our method with other state-of-the-arts on THUMOS14 dataset for action localization, including both full supervision and weak supervision.}
		\label{table:headings}
		\begin{tabular}{m{1.5cm}<{\centering}m{2.5cm}m{0.5cm}<{\centering}m{0.5cm}<{\centering}m{0.5cm}<{\centering}m{0.5cm}<{\centering}m{0.5cm}<{\centering}m{0.5cm}<{\centering}m{0.5cm}<{\centering}}
			\toprule \noalign{\smallskip} 
			\multirow{2}{1.5cm}{Supervision} & \multirow{2}{2.5cm}{Method}  & \multicolumn{7}{c}{mAP@\textit{tIoU} ($\alpha $)}\\
			& & 0.1 & 0.2 & 0.3 & 0.4 & 0.5 & 0.6 & 0.7\\
			\noalign{\smallskip}
			\hline
			\noalign{\smallskip}
			\multirow{5}{1.5cm}{Fully supervised} 
			& Oneata et al. \cite{D.Oneata} & 36.6 & 33.6 & 27.0 & 20.8 & 14.4 & - & -\\
			& Richard et al. \cite{Richard} & 39.7 & 35.7 & 30.0 & 23.2 & 15.2 & - & -\\
			& Shou et al. \cite{Z.Shou} & 47.7 & 43.5 & 36.3 & 28.7 & 19.0 & 10.3 & 5.3\\
			& Yuan et al. \cite{Yuan} & 51.0 & 45.2 & 36.5 & 27.8 & 17.8 & - & -\\
			& Lin et al. \cite{T.Lin} & 50.1 & 47.8 & 43.0 & 35.0 & 24.6 & 15.3 & 7.7\\
			& Zhao et al. \cite{Y.Zhao} & 60.3 & 56.2 & 50.6 & 40.8 & 29.1 & - & -\\
			& Gao et al. \cite{Gao} & 60.1 & 56.7 & 50.1 & 41.3 & \textbf{31.0} & 19.1 & 9.9\\
			\hline
			\noalign{\smallskip}
			\multirow{4}{1.5cm}{Weakly supervised} & Singh et al \cite{Singh}  & 36.4 & 27.8 & 19.5 & 12.7 &6.8 & - & -\\
			& Wang et al \cite{L.Wang}  & 44.4 & 37.7 & 28.2 & 21.1 & 13.7 & - & - \\ 
			& \multirow{1}{2.5cm}{CPMN}  & \textbf{47.1} & \textbf{41.6} & \textbf{32.8} & \textbf{24.7} & \textbf{16.1} & 10.1 & 5.5\\
			\bottomrule
		\end{tabular}
	\end{center}
	\vspace{-0.1em}
\end{table}
\setlength{\tabcolsep}{1.5pt}

\renewcommand{\multirowsetup}{\centering}
\setlength{\tabcolsep}{7pt}
\begin{table}[t] 
	\begin{center}
		\caption{Results on validation set of ActivityNet-1.3 in terms of mAP@\textit{tIoU} and average mAP. Note that all compared methods are fully supervised. }
		\label{table:headings}
		\begin{tabular}{m{2.7cm}<{\centering}m{2.6cm}m{0.8cm}<{\centering}m{0.8cm}<{\centering}m{0.8cm}<{\centering}m{1cm}<{\centering}}
			\toprule \noalign{\smallskip} 
			Supervision &\multirow{1}{2.6cm}{Method} & 0.5 & 0.75 & 0.95 & Average\\
			\noalign{\smallskip}
			\hline
			\noalign{\smallskip}
			\multirow{5}{2.7cm}{Fully supervised} &
			Singh et al. \cite{G.Singh} & 34.5 & - & - & - \\ 
			&Heilbron et al. \cite{SCC} & 40.00 & 17.90 & 4.70 & 21.70 \\
			&Wang et al. \cite{Uts} & 42.28 & 3.76 & 0.05 & 14.85 \\
			&Shou et al. \cite{CDC} & 43.83 & 25.88 & 0.21 & 22.77 \\
			&Xiong et al. \cite{Y.Xiong} & 39.12 & 23.48 & 5.49 & 23.98 \\
			&Lin et al. \cite{TCAP} & 48.99 & 32.91 & 7.87 & 32.26 \\
			\hline
			\noalign{\smallskip}
			\multirow{1}{2.7cm}{Weakly supervised} &
			\multirow{1}{2.6cm}{CPMN}
			& \multirow{1}{0.8cm}{39.29} & \multirow{1}{0.8cm}{24.09} & \multirow{1}{0.8cm}{6.71} & \multirow{1}{1cm}{24.42} \\
			\bottomrule
		\end{tabular}
	\end{center}
	\vspace{-0.4em}
\end{table}
\setlength{\tabcolsep}{1.5pt}

Fig. 6 visualizes the localization performance of our proposed method on THUMOS14 dataset. As shown in Fig. 6 (a), the video includes two different action classes, and our Attention-based Cascaded (AC) T-CAS of corresponding class can localize the entire regions of target actions individually. It confirms that the two cascaded classifiers are successful in mining different but complementary target regions. And in Fig. 6 (b), there are many action instances densely distributed in the video, however, our method can effectively highlight the regions which may contain actions, which demonstrates that our modified T-CAS is able to generate dense detections. Besides, the Fig. 6 (c) presents a video with similar appearance and little dynamic motions along the temporal dimension, which is a difficult scene even for humans to distinguish between adjacent frames, resulting in inevitably missing detections. Nevertheless, our algorithm is still robust enough to discover some discriminative regions even with some false positives. To conclude, the proposed cascaded action localization sequence together with the pyramidal attention map is intuitive to promote the overall localization performance.

\begin{figure}[t]
	\setlength{\abovecaptionskip}{-0.2cm}  
	\setlength{\belowcaptionskip}{-0.6cm} 
	\begin{center}
		\begin{minipage}[b]{1.0\linewidth}
			\centering
			\centerline{\includegraphics[height=4.3cm]{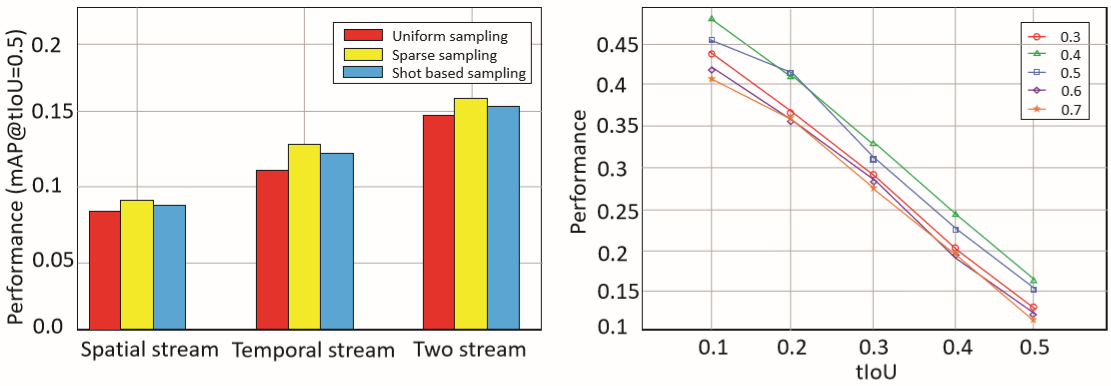}}
			\medskip
		\end{minipage}
	\end{center}
	\caption{Evaluation of the CPMN with different sampling methods \textbf{(left)} and erasing thresholds \textbf{(right)} used in the CCM on THUMOS14.}
	\label{fig_erasing}
\end{figure}

\subsection{Ablation Study}
In this section, we evaluate CPMN with different implementation variations to study their effects and investigate the contribution of several components proposed in our network. All the experiments for our ablation study are conducted on THUMOS14 dataset.

\textbf{Sampling strategy.} The input untrimmed video usually exists substantial redundant frames which are useless for the cascaded model to leverage discriminative regions for recognition. In order to reduce computational cost, we sample the input feature sequence instead of using all units for video categorization. We first evaluate the Cascaded Classification Module (CCM) with different sampling methods, including uniform sampling, sparse sampling and shot-based sampling. The evaluation results are illustrated in Fig. 5 (left). We can observe that the shot-based sampling method which takes action structure into consideration shows better performance than uniform sampling and sparse sampling which serves as a data augmentation step leads to the best performance. Therefore we finally adopt sparse sampling to sample the input feature sequence of the CCM during training phase.

\textbf{Erasing threshold.} We continue to study the influence of different erasing thresholds $ \zeta $ which we use to identify the discriminative regions highlighted by the first classifier of the CCM and create a mask for online adversarial erasing step. As shown in Fig. 5 (right), we test the thresholds from 0.3 to 0.7 and report the performance over different IoU. We observe that when the threshold $ \zeta=0.4 $, the localization performance of the two-stage model is boosted and the value larger or smaller than 0.4 would fail to encourage the second classifier to mine entire regions of target actions and may bring background noise.

\textbf{Architecture of CPMN.} As shown in Table 3, we investigate the contribution of each module proposed in our method. We choose the architecture sharing the same idea with CAM \cite{B.Zhou} for discriminative localization as our baseline model without online erasing step to mine entire regions and the pyramidal attention map. The comparison results reveal that the original T-CAS together with the mining T-CAS can promote the performance and with the help of pyramidal attention map, the localization performance can be further boosted. These observations suggest that these modules are all effective and indispensable in CPMN. Note that we also test the CCM with more classifiers. Specifically, we add the third classifier to handle the erased feature maps guided by the first two classifiers. However, we don't find any significant improvement.

\setlength{\tabcolsep}{15pt}
\begin{table}[t]
	\setlength{\belowcaptionskip}{0.3cm}  
	\begin{center}
		\caption{Performance with respect to architecture choices. In first column, only original T-CAS without adversarial mining and attention map is used for action localization.}
		\label{table:headings}
		\begin{tabular}{lc<{\centering}c<{\centering}c<{\centering}}
			\toprule \noalign{\smallskip}
			Original T-CAS &  \checkmark    &\checkmark &\checkmark\\
			Mining T-CAS  &     &\checkmark &\checkmark\\
			Pyramidal Attention Map (PAM) &     &  &\checkmark\\
			\noalign{\smallskip}
			\hline
			\noalign{\smallskip}
			mAP ($\alpha $ = 0.5) & 11.4  & 14.5 & 16.1\\
			\bottomrule
		\end{tabular}
	\end{center}
	\vspace{-0.6em}
\end{table}

\setlength{\tabcolsep}{1.4pt}
\begin{figure}
	\setlength{\abovecaptionskip}{-0.4cm}  
	\setlength{\belowcaptionskip}{-0.3cm} 
	\begin{center}
		\begin{minipage}[b]{1.0\linewidth}
			\centering
			\centerline{\includegraphics[height=7.6cm]{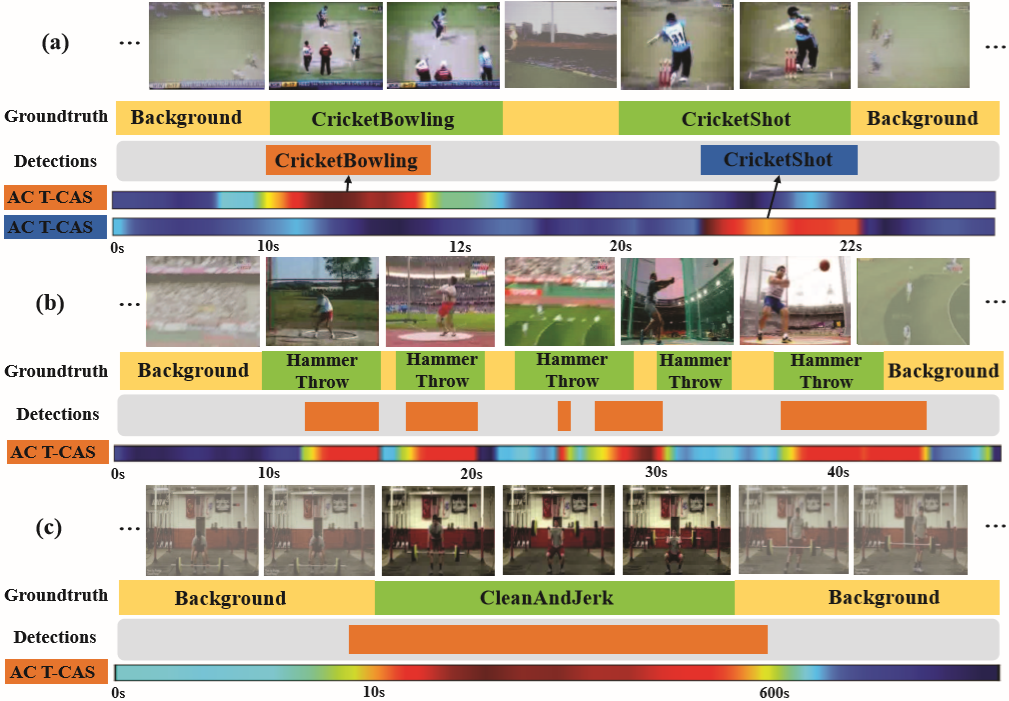}}
			\medskip
		\end{minipage}
	\end{center}
	\caption{Visualization of the action instances located by CPMN on THUMOS14. Figure (a) shows that entire regions of two actions are separately located in corresponding T-CAS. Figure (b) shows that dense predictions can be generated by our approach with less missing detections. Figure (c) shows that even the instances with small length and similar appearance, our model still can locate it with less false positives.}
	\label{fig_qualitative}
\end{figure}

\section{Conclusion}
In this paper, we propose the Cascaded Pyramid Mining Network (CPMN) for weakly supervised temporal action localization. Our method includes two main steps: cascaded adversarial mining and pyramidal attention map inference. Cascaded adversarial mining is realized by designing two cascaded classifiers to collaborate on locating entire regions of target actions with a novel online adversarial erasing step. And the pyramidal attention map tries explicitly handling the falsely activated regions and missing detections in the localization sequence, which is inferred upon the prediction results of the multi-tower network. Extensive experiments reveal that CPMN significantly outperforms other state-of-the-art approaches on both THUMOS14 and ActivityNet-1.3 datasets.

\clearpage
%
%
%
\bibliographystyle{splncs04}
\bibliographystyle{unsrt}
\bibliography{ref}
\end{document}